\documentclass[10pt,twocolumn,letterpaper]{article}

\usepackage{ifpdf} 
\usepackage{iccv}
\usepackage{times}
\usepackage{graphicx}
\usepackage{amsmath}
\usepackage{amssymb}
\usepackage{amsthm}
\usepackage{subcaption}
\usepackage{booktabs}
\usepackage{multirow}
\usepackage{colortbl}
\definecolor{mygray}{gray}{.9}

\usepackage[pagebackref=true,breaklinks=true,letterpaper=true,colorlinks,bookmarks=false]{hyperref}

\iccvfinalcopy 

\def\iccvPaperID{3213} 

\begin{document}

\title{Correlation Maximized Structural Similarity Loss for Semantic Segmentation}

\author{Shuai Zhao$^1$,
Boxi Wu$^1$,
Wenqing Chu$^1$,
Yao Hu$^2$,
and Deng Cai$^{1}$\thanks{Deng Cai is the corresponding author} \\
$^1$State Key Lab of CAD\&CG, College of Computer Science, Zhejiang University, China\\
$^2$Alibaba Youku Cognitive and Intelligent Lab \\
{\tt\small \{zhaoshuaimcc, wqchu16\}@gmail.com},
{\tt\small wuboxi@zju.edu.cn},
{\tt\small yaoohu@alibaba-inc.com},
{\tt\small dcai@zju.edu.cn}
}


\maketitle

\begin{abstract}
Most semantic segmentation models treat semantic segmentation as a pixel-wise
classification task and use a pixel-wise classification error as their optimization criterions.
However, the pixel-wise error ignores the strong dependencies among
the pixels in an image, which limits the performance of the model.
Several ways to incorporate the structure information of the objects have been investigated,
\eg, conditional random fields (CRF), image structure priors based methods,
and generative adversarial network (GAN).
Nevertheless, these methods usually require extra model branches or additional memories,
and some of them show limited improvements.
In contrast, we propose a simple yet effective structural similarity loss (SSL)
to encode the structure information of the objects,
which only requires a few additional computational resources in the training phase.
Inspired by the widely-used structural similarity (SSIM) index in image quality assessment,
we use the linear correlation between two images to quantify their structural similarity.
And the goal of the proposed SSL is to pay more attention to the positions,
whose associated predictions lead to a low degree of linear correlation between two corresponding regions
in the ground truth map and the predicted map.
Thus the model can achieve a strong structural similarity
between the two maps through minimizing the SSL over the whole map.
The experimental results demonstrate that our method can achieve
substantial and consistent improvements in performance
on the PASCAL VOC 2012 and Cityscapes datasets.
The code will be released soon.
\end{abstract}

\section{Introduction} \label{01_introduction}
\begin{figure}[ht]
	\centering
	\begin{subfigure}[]{0.22\textwidth}
		\centering
		\includegraphics[width=0.95\textwidth]{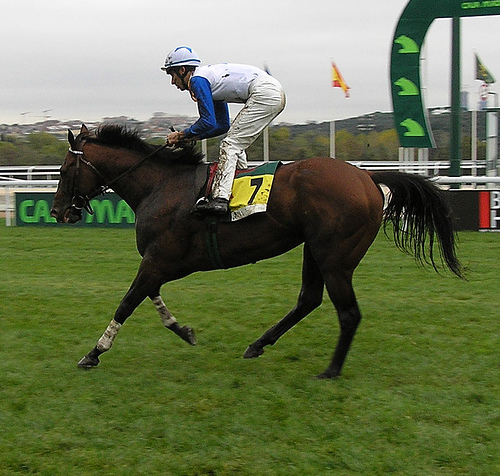}
		\caption{{\footnotesize Real-world image} }
	\end{subfigure}
	\begin{subfigure}{0.22\textwidth}
		\centering
		\includegraphics[width=0.95\textwidth]{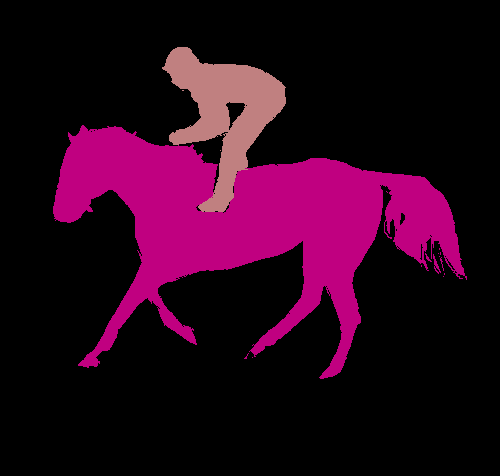}
		\caption{{\footnotesize Ground truth}}
	\end{subfigure} \\
	\begin{subfigure}{0.22\textwidth}
		\centering		
		\includegraphics[width=0.95\textwidth]{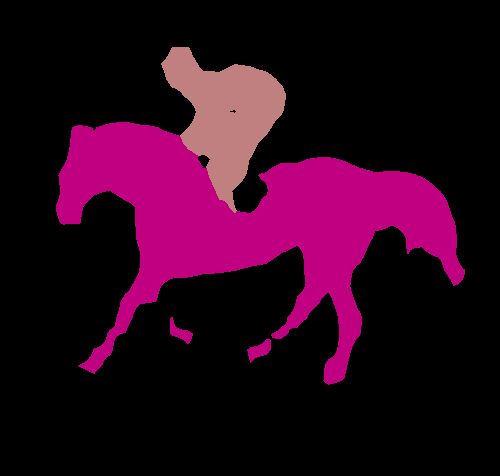}
		\caption{{\footnotesize Prediction with cross entropy}}
	\end{subfigure}
	\begin{subfigure}{0.22\textwidth}
		\centering
		\includegraphics[width=0.95\textwidth]{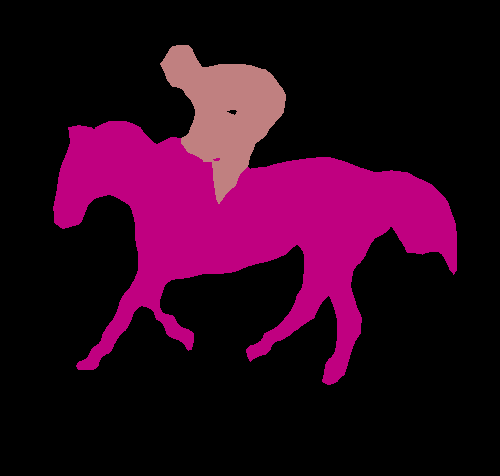}
		\caption{{\footnotesize Prediction with proposed SSL} }
	\end{subfigure}
	\caption{
		Compared to the regular cross entropy loss, 
		the SSL can incorporate the structure information of the objects.		
		Thus the model trained with SSL can better identify
		the pixels belonging to the horse's leg or the horse body covered by the race number.
	}
	\label{fig_001}
\end{figure}
Semantic image segmentation is considered as
a pixel-wise multiclass classification problem in practice,
and the goal is to assign semantic labels to every pixel in the image.
Much progress has been made with powerful convolutional neural networks 
(\eg, VGGNet~\cite{2014_VGGNet}, ResNet~\cite{2016_ResNet}, Xception~\cite{2017_Xception})
and fancy segmentation models
(\eg, FCN~\cite{2015_CVPR_FCN}, PSPNet~\cite{2017_CVPR_PSPNet},
DeepLab~\cite{chenDeepLabv2,chenDeepLabv3,chenDeepLabv3plus}, ExFuse~\cite{2018_ExFuse}).
Generally, these models are optimized by a pixel-wise classification error.
And the most commonly used pixel-wise loss function
for semantic segmentation is the softmax cross entropy loss:
\begin{align}
\mathcal{L}_{ce}(y, p) = -\frac{1}{N} \sum_{n=1}^{N} \sum_{c=1}^{C} y_{n,c}\log(p_{n,c}), \label{equ_001}
\end{align}
where $N$ is the number of pixels, $C$ is the number of classes, 
$y \in \{0,1\}$ is the ground truth, and $p \in [0, 1]$ is the estimated probability.
The pixel-wise cross entropy loss considers the pixels as independent samples,
and the total loss used for training is the average loss over all pixels.
However, there are strong dependencies existed among pixels in an image and these    
dependencies carry important information about the structure of the objects~\cite{2004_SSIM}.
As the pixel-wise loss ignores the relationship between pixels,  
models trained with a pixel-wise loss
may get poor segmentation results when the visual evidence for the foreground
is weak (\eg, horse body covered by the race number in Fig.~\ref{fig_001})
or when pixels belong to objects with small spatial structure
(\eg, pixels belonging to the horse's leg in Fig.~\ref{fig_001})~\cite{2018_ECCV_AAF}.

Previous work has pursued three directions to utilize the structure information of the objects:
\begin{enumerate}
	\item \textbf{Conditional Random Field (CRF)}.
	CRF can model the relationships between the pixels 
	and enforce the predictions of pixels with similar visual appearances to be more consistent.
	It is typically used as a post-processing step~\cite{chenDeepLabv2,2011_Efficient_CRF,2017_CVPR_Shen} or a plug-in module inside 
	the neural networks~\cite{2018_Ziwei,2015_ICCV_Zheng}.
	Nevertheless, CRF usually has time-consuming iterative inference routines and is sensitive to visual appearance changes~\cite{2018_ECCV_AAF}.
	
	\item \textbf{Image structure priors}, such as contour cues~\cite{2016_CVPR_Bertasius,2016_CVPR_Chen_Edge} and pixel affinity~\cite{2018_ECCV_AAF,2017_NIPS_Liu,2016_CVPR_Affinity_CNN}.
	Some additional model branches in such approaches~\cite{2016_CVPR_Bertasius,2016_CVPR_Chen_Edge, 2016_CVPR_Affinity_CNN} are required to extract contours or pixel affinity from images.
	Then these image priors are fused into predicted maps in a carefully designed manner.
	Besides, additional memory is required to hold the pixel affinity matrix~\cite{2018_ECCV_AAF, 2017_NIPS_Liu}.
	
	\item \textbf{Generative adversarial network (GAN)}~\cite{2017_Isola,2016_pauline, 2017_ICCV_CycleGAN}.
	Luc \etal~\cite{2016_pauline} think the discriminator
	can detect and correct high-order inconsistencies between the ground truth map and the one produced by the generator.
	However, GAN models are often hard to train, and the performance of them
	may become unstable or even collapse~\cite{2015_ALec}.
	Additionally, a large memory is also required to hold deep generator and discriminator networks simultaneously.
	Even with such efforts, the improvement of the performance may be slight,
	\eg, Luc \etal~\cite{2016_pauline} get about 0.25\% improvement in mean intersection-over-union
	(mIoU) score on the PASCAL VOC 2012 dataset~\cite{2010_PASCAL}. 
\end{enumerate}
As discussed above, most of the existing approaches are 
resource-consuming or show minor improvements.
Consequently, the top-performing models (\eg, DeepLabv3+~\cite{chenDeepLabv3plus}, MSCI~\cite{2018_Lin_ECCV}, ExFuse~\cite{2018_ExFuse}) typically do not integrate these structure modeling techniques.
In contrast, we propose a simple yet effective novel method to encode the 
structure information of the objects,
which only requires a few additional computational resources during the training stage.

Inspired by the widely-used structural similarity (SSIM) index~\cite{2004_SSIM}
in image quality assessment (IQA),
we use the linear correlation between two images to quantify their structural similarity.
Then we calculate the differences between the standard normalized results of two corresponding regions
in the ground truth map and the predicted map.
The differences between the two regions are used to measure their degree of linear correlation,
and it is considered as the structural difference between the two regions.
According to the magnitude of the structural difference,
the regular cross entropy loss is reweighted to pay more attention to the
inconsistent pixels between the two regions.
Meanwhile, pixels with small structural differences are
abandoned to make the training more efficient,
and this can also be regarded as the online hard example mining
(OHEM) ~\cite{2017_CVPR_Full_RR,2016_CVPR_OHEM,1996_bootstrap, 2016_bootstrap} strategy.
Now we get our structural similarity loss (SSL) for semantic segmentation.
With the SSL, the segmentation model learns to focus more on positions
whose relevant predictions lead to a low degree of linear correlation
between two corresponding regions in two maps.
By minimizing the SSL over the whole map,
the predicted map can achieve a strong structural similarity with the ground truth map.
And it is important to note that the SSL is supervised by the statistics of regions in the maps,
thus it is a region-wise loss rather than a pixel-wise loss.

The SSL has a few appealing properties over existing approaches.
First, the loss provides an intuitive way to measure the
structural similarity between two images.
Second, it can be implemented easily in a convolutional manner
and only requires a few additional computational resources during training.
Thus it can be effortlessly incorporated into any segmentation frameworks.
Last but not least, the SSL is also easier to train than GAN and more efficient than CRF,
as the SSL does not require additional networks or inference routines during training and testing.

The experimental results demonstrate that our proposed method can achieve
substantial and consistent improvements in performance
on the PASCAL VOC 2012~\cite{2010_PASCAL} and Cityscapes~\cite{2016_Cityscapes} datasets.
We also empirically compare the SSL with some existing structure modeling methods~\cite{2018_ECCV_AAF,2011_Efficient_CRF} on the PASCAL VOC dataset,
where the SSL obtains better results.

\section{Related Work}
\subsection{Semantic segmentation}
Most state-of-the-art semantic segmentation approaches
(\eg, PSPNet~\cite{2017_CVPR_PSPNet}, DeepLabv3~\cite{chenDeepLabv3},
DeepLabv3+~\cite{chenDeepLabv3plus}, SDN~\cite{2017_SDN},
EncNet~\cite{2018_CVPR_EncNet}, ExFuse~\cite{2018_ExFuse})
use a pixel-wise error as their optimization criterions.
However, the pixel-wise error ignores the relationship between pixels.
As discussed in Sec.~\ref{01_introduction},
several ways to incorporate the structure information of the objects have been investigated,
\eg, CRF based methods~\cite{chenDeepLabv2,2011_Efficient_CRF,2017_CVPR_Shen,2018_Ziwei,2015_ICCV_Zheng},
image structure priors based methods~\cite{2016_CVPR_Bertasius,2016_CVPR_Chen_Edge, 2018_ECCV_AAF,2017_NIPS_Liu,2016_CVPR_Affinity_CNN},
and GAN~\cite{2017_Isola, 2016_pauline, 2017_ICCV_CycleGAN}.
Nevertheless, these methods are usually resource-consuming or show minor improvements.

\subsection{Structural similarity index}
The SSIM index~\cite{2004_SSIM} is a widely used full-reference IQA measure.
The goal of the full-reference IQA algorithm is to measure the similarity between two images,
where one of them is the reference image.
For semantic segmentation, the ground truth map is the reference image,
and we need to measure the similarity between the ground truth map and the predicted map.
Given two images or two image patches $\mathbf{x}$ and $\mathbf{y}$, the SSIM index combines three components, namely, a luminance term, a contrast term and a structure term as follows:
\begin{align}
l(\mathbf{x}, \mathbf{y}) &= \frac{2 \mu_x \mu_y + C_1}{\mu_x^2 + \mu_y^2 + C_1}, \label{equ_002}\\
c(\mathbf{x}, \mathbf{y}) &= \frac{2 \sigma_x \sigma_y + C_2}{\sigma_x^2 + \sigma_y^2 + C_2}, \\
s(\mathbf{x}, \mathbf{y}) &= \frac{\sigma_{xy} + C_3}{\sigma_x \sigma_y + C_3},
\label{equ_004}
\end{align}  
where $\mu_x$, $\sigma_x^2$ and $\sigma_{xy}$ are the mean of $\mathbf{x}$, the variance of $\mathbf{x}$, 
and the covariance of $\mathbf{x}$ and  $\mathbf{y}$ respectively. The small positive constants 
$C_1$, $C_2$ and $ C_3$ are included to stabilize each term. The mean and variance 
can be considered as estimates of the lumiance and contrast of the image, and the covariance $\sigma_{xy}$
measures the tendency of  $\mathbf{x}$ and  $\mathbf{y}$ to vary together, 
thus an indication of structural similarity~\cite{2004_MSSSIM}. 

Generally, the SSIM index is expressed as 
\begin{align}
\mathrm{SSIM}(\mathbf{x}, \mathbf{y}) = [l(\mathbf{x}, \mathbf{y})]^\alpha [c(\mathbf{x}, \mathbf{y})]^\theta
[s(\mathbf{x}, \mathbf{y})]^\gamma,
\end{align}
where parameters $\alpha >0$,  $\theta >0$ and $\gamma > 0$ are used to adjust the relative importance of the three components.
If we set $\alpha = \theta = \gamma = 1$ and $C_3 = C_2 / 2$,
we get the commonly used simplified form of the SSIM index~\cite{2009_love_or_leave, 2004_SSIM, 2004_MSSSIM}:
\begin{align}
\mathrm{SSIM}(\mathbf{x}, \mathbf{y}) &= \Big( \frac{2 \mu_x \mu_y + C_1}{\mu_x^2 + \mu_y^2 + C_1} \Big)
\Big( \frac{2 \sigma_{xy} + C_2}{\sigma_x^2 + \sigma_y^2 + C_2} \Big),  \nonumber \\
&= S_1(\mathbf{x}, \mathbf{y}) S_2(\mathbf{x}, \mathbf{y}). \label{equ_ssim_lucs}
\end{align}
Here $-1 \le \mathrm{SSIM}(\mathbf{x}, \mathbf{y}) \le 1$ 
and $\mathrm{SSIM}(\mathbf{x}, \mathbf{y})=1$ if and only if $\mathbf{x}=\mathbf{y}$~\cite{2004_SSIM}. 
$S_2(\mathbf{x}, \mathbf{y})$ is also called contrast-structure measure.

\noindent
\textbf{SSIM-based optimization.} The SSIM index is also widely used as an optimization criterion in 
some image processing tasks for the sake of better visual quality,
\eg, image denosing~\cite{2008_Stat_SSIM,2006_A_Linear,2015_loss_for_image_restoration},
image downscaling~\cite{2015_PerceptuallyBD},
image deblurring~\cite{2014_solving}, image compression~\cite{2009_love_or_leave},
image demosaicking and super-resolution~\cite{2015_loss_for_image_restoration}.
Conventionally, the maximization of SSIM is cast as a minimization problem:
\begin{align}
\mathcal{L}_{ssim}(\mathbf{x}, \mathbf{y}) = 1 - \mathrm{SSIM}(\mathbf{x}, \mathbf{y}) \label{equ_ssim_loss_lucs}.
\end{align} 
The SSIM-based optimization is an instance of the perceptual optimization framework
where the objective measure models the perceptual quality of an image~\cite{2018_Brunet}.

\section{Methods}
In this section, we first analyze the SSIM index.
Then we propose our structural similarity loss for semantic segmentation.
Finally, we discuss the estimation of the local statistic
and the overall objective function used for training.

\subsection{Analysis of the SSIM index} \label{sec_3_1}
The structure term (Eq.~\eqref{equ_004}) is the key that the SSIM index
can measure the structural similarity between two images.
The term is exactly the pearson correlation coefficient between $\mathbf{x}$ and $\mathbf{y}$:
\begin{align}
\rho_{\mathbf{x}, \mathbf{y}} = \frac{\mathbb{E}\big[ (\mathbf{x} - \mu_x)(\mathbf{y} - \mu_y ) \big] }{\sigma_x \sigma_y}.
\end{align}
This suggests that the SSIM index uses the linear correlation between two images
to quantify their structural similarity.
Image processing systems~\cite{2008_Stat_SSIM,2014_solving,2015_PerceptuallyBD,
2009_love_or_leave,2015_loss_for_image_restoration},
which use the $\mathcal{L}_{ssim}$ as an optimization criterion for better visual quality,
are actually attempting to achieve a high positive linear correlation
between the generated image and the ground truth image.
However, as the $\mathcal{L}_{ssim}$ is not convex~\cite{2012_Brunet} and
the segmentation map is also not the same as the real-world image,
the $\mathcal{L}_{ssim}$ is not an appropriate optimization criterion for semantic segmentation.

We can get another simplified form of the $\mathcal{L}_{ssim}$
by subtracting the mean from the values of the pixels~\cite{2012_Brunet,2006_A_Linear}:
\begin{align}
\mathcal{L}_{ssim}(\mathbf{x}_m, \mathbf{y}_m) &= 1 - S_2(\mathbf{x}_m, \mathbf{y}_m) \nonumber \\
&= \frac{\sigma_x^2 + \sigma_y^2 - 2 \sigma_{xy}}{\sigma_x^2 + \sigma_y^2 + C_2} \nonumber \\ 
&=\frac{\Vert\mathbf{x}_m - \mathbf{y}_m \Vert^2}{\Vert \mathbf{x}_m \Vert^2 + \Vert \mathbf{y}_m \Vert^2 +(N-1)C_2}, \label{equ_ssim_loss_cs}
\end{align}
where $\sigma_x^2 = \frac{1}{N-1}\Vert\mathbf{x} - \mu_x \Vert^2$ and
$N$ is the number of entries of $\mathbf{x}$.
Here $\mathbf{x}_m = \mathbf{x} - \mu_x$, $\mathbf{y}_m = \mathbf{y} - \mu_y$, and
it means $S_1(\mathbf{x}_m, \mathbf{y}_m)=1$ because $\mu_{\mathbf{x}_m}=0$ and $\mu_{\mathbf{y}_m}=0$.
The equation~\eqref{equ_ssim_loss_cs} is quasiconvex when
$\mathbf{x}_m^T\mathbf{y}_m \ge \big(-(N-1)C_2/2 \big)$~\cite{2012_Brunet},
but this condition is not guaranteed to be met in the context of semantic segmentation.
As the ground truth is in $\{0, 1\}$ and the predicted probability is in $[0, 1]$,
we can only know the elements of $\mathbf{x}_m$ and $\mathbf{y}_m$ are in range $(-1, 1]$.
Moreover, quasiconvex optimization requires carefully designed optimization methods like
bisection method~\cite{2004_Boyd, 2008_Stat_SSIM}.
For these reasons,  the equation~\eqref{equ_ssim_loss_cs} is also not an appropriate 
optimization criterion for a deep neural network semantic segmentation model.

\subsection{Structural similarity loss}

Based on the analysis in Sec.~\ref{sec_3_1}, we propose our structural similarity loss (SSL)
for semantic segmentation to achieve a high positive linear correlation between
the ground truth map and the predicted map.

For a ground truth map with shape $H \times W \times C$
($H, W, C$ are the height, width and the number of channels respectively),
we consider it as $C$ binary images.
The structure comparison is conducted after standard normalization:
\begin{align}
	e =\Big \vert \frac{y - \mu_y + C_4}{ \sigma_y + C_4} - \frac{p -  \mu_p+ C_4}{\sigma_p + C_4} \Big \vert,
\end{align}
where $\mu_y$ and $\sigma_y$ is the \textit{local} mean and standard deviation of the ground truth $y$ respectively, 
the corresponding point of $y$ locates at the center of the local region,
$p$ is the predicted probability,
and $C_4=0.01$ is a stability factor.
The total absolute error $e$ between two image patches can measure
their degree of linear correlation.
The smaller the total error $e$ is, the more likely the two image patches achieve a positive linear correlation,
and this means the structures of them are more likely to be the same.

Then we reweight the cross entropy loss according to $e$
and abandon the samples with small $e$:
\begin{align}
	f_{n,c} &= \mathbf{1}\{ e_{n,c} > \beta e_{max} \},  \label{equ_fnc} \\ 
	\mathcal{L}_{ssl}(y_{n,c}, p_{n,c}) &=  e_{n,c} f_{n,c} \mathcal{L}_{ce}(y_{n,c}, p_{n,c}),
%
\end{align}
where $e_{max}$ is the theoretical maximum value of $e$,
$\mathbf{1}\{\cdot\}$ equals one when the condition inside holds and otherwise equals zero,
$\beta \in [0, 1)$ is a weight factor used to select the abandoned samples,
and $\mathcal{L}_{ce}$ is the sigmoid cross entropy loss.
The factor $\beta$ is set to be $0.1$ in practice, and this is an empirical value.

\begin{figure}[t]
	\centering
	\begin{subfigure}[]{0.2\textwidth}
		\centering
		\includegraphics[width=0.8\textwidth]{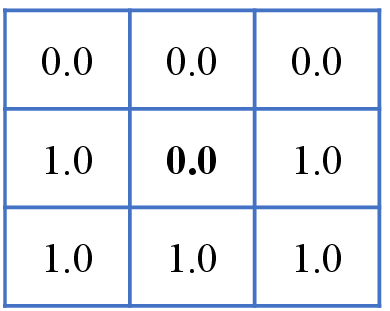}
		\caption{Ground truth}
	\end{subfigure}
	\begin{subfigure}{0.2\textwidth}
		\centering
		\includegraphics[width=0.8\textwidth]{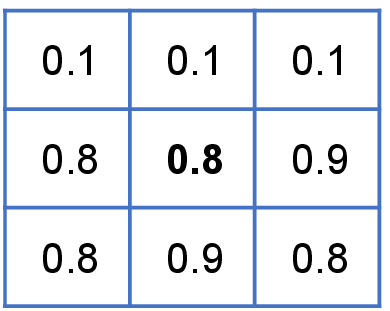}
		\caption{Prediction}
	\end{subfigure} \\
	\begin{subfigure}{0.2\textwidth}
		\centering
		\includegraphics[width=0.8\textwidth]{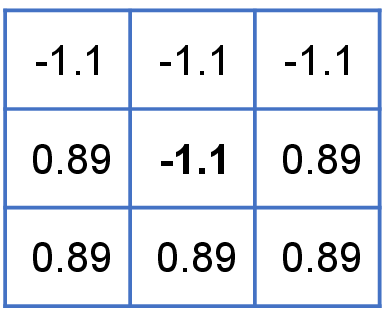}
		\caption{Normalized groun truth}
	\end{subfigure}
	\begin{subfigure}{0.2\textwidth}
		\centering
		\includegraphics[width=0.8\textwidth]{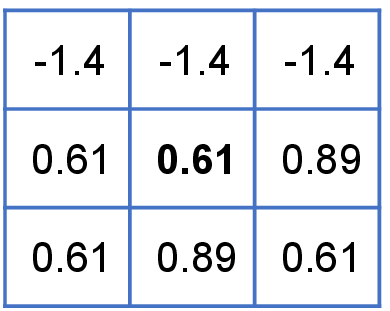}
		\caption{Normalized prediction}
	\end{subfigure}
	\caption{
		The total binary cross entropy loss between (a) and (b) is about $2.805$,
		and the loss of the center pixel accounts for about $\mathbf{57}\%$ of the total loss. 
		After normalization and reweighting,
		the total loss between (c) and (d) is about $3.060$,
		and the loss of the center pixel accounts for about $\mathbf{91}\%$ of the total loss.
		The inconsistent pixels between two maps get more attention after normalization and reweighting.
	}
	\label{fig_002}
\end{figure}

The influence of the reweighting strategy is shown in Fig.~\ref{fig_002}.
The inconsistent pixels between two maps get more attention after reweighting.
As the maps are standardly normalized locally,
the SSL is under the supervision of the local statistics,
thus it is a region-wise loss.
Moreover, Janocha \etal~\cite{2017_on_loss} experimentally demonstrate that the log
loss is a very appropriate choice when training a deep neural network classifier,
so we still use it as the optimization criterion.
And the error $e$ is used as a constant weighting coefficient
to make the model easy to optimize in practice.

Generally, there are millions or even tens of millions of pixels in a mini-batch.
At the late stage of the training, the segmentation model can usually get a
high pixel accuracy (\eg, 96\%) and a relatively low mean intersection-over-union (mIoU) score (\eg, 78\%). 
This phenomenon indicates that the easily classified samples dominate the loss
and make the training inefficient~\cite{2017_focal,2017_CVPR_DLC}.
Therefore, we consider the samples with small $e$ as easy examples~\cite{2016_CVPR_OHEM} and
abandon them during training.
Then the hard examples (the error $e$ is large)
which leads to a low degree of linear correlation between $y$ and $p$ further get more attention.
This is the OHEM strategy
~\cite{2017_CVPR_Full_RR,2016_CVPR_OHEM,1996_bootstrap, 2016_bootstrap}.

The total SSL over the mini-batch is:
\begin{align}
	\mathcal{L}_{ssl}(y, p) =  \frac{1}{M} \sum_{n=1}^{N} \sum_{c=1}^C \mathcal{L}_{ssl}(y_{n,c}, p_{n,c}), \label{equ_012}
\end{align}
where $M = \sum_{n=1}^N \sum_{c=1}^C f_{n,c}$ is the number of hard examples.
As the structure comparison is performed between each ground truth binary image
and the corresponding probability map independently,
we consider each point in the binary image as a sample
and the points in different binary images are independent.
These factors determine the manner we calculate the number of hard examples and the choice
of the sigmoid cross entropy loss.

With the SSL, the segmentation model learns to pay more attention to
predictions which lead to a low degree of linear correlation between two regions.
Thus the prediction $p$ can achieve a strong structural similarity with the ground truth
$y$ by minimizing the SSL over the whole map.

\subsection{Estimation of the local statistic}
The local statistics $\mu_y$ and $\sigma_y$ are computed
within a local square window with certain size $k$,
which moves pixel-by-pixel over the entire image.
This can be easily implemented in a convolutional manner.
Following the SSIM index~\cite{2004_SSIM}, we use a circular-symmetric
gaussian weighting function $\mathbf{w}=\{w_i | i=1,2, \ldots ,k^2\}$  
(normalized to unit sum $\sum_{i=1}^{k^2} w_i = 1$) with standard deviation of 1.5 samples to estimate
the local statistics:
\begin{align}
	\mu_y &= \sum_{i=1}^{k^2} w_i y_i, \\
	\sigma_y^2 &= \sum_{i=1}^{k^2}w_i(y_i - \mu_y)^2 = \sum_{i=1}^{k^2}w_iy_i^2 - \mu_y^2. \label{equ_014}
\end{align}
With the gaussian window approach,
the normalized segmentation map exhibit a locally isotropic property~\cite{2004_SSIM}.  
And in the gaussian window, points closer to the center will contribute more information.
As the ground truth $y$ is in $\{0, 1\}$, it means $y^2 = y$.
If we subtitute it into the Eq.~\eqref{equ_014}, we get $\sigma_y^2 = \mu_y - \mu_y^2$.
Then we can get:
\begin{align}
		y^{nor} = \frac{y - \mu_y + C_4}{ \sqrt{\mu_y - \mu_y^2} + C_4}.
\end{align}
\begin{figure}[t]
	\centering
	\includegraphics[width=0.46\textwidth]{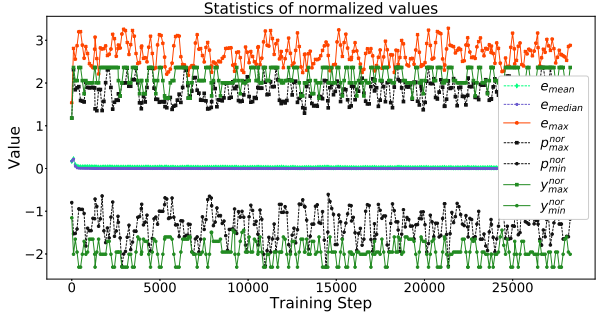}
	\caption{Statistics of normalized values during a training procedure.
		The standard deviation of the gaussian kernel is 1.5, and the region size is 3.
		The data is smoothed to get better visual quality.
		The black line and dark green line represent the extreme value of $p^{nor}$
		and $y^{nor}$ respectively.
		It is clear that the extreme value of $\vert p^{nor} \vert$ is
		less than the extreme value of $\vert y^{nor} \vert $ at the same time.
		The uppermost line represents the maximum practical value of $e$ during training.
		The middle lines represent the mean and median of $e$.
		They are both close to 0 and the mean is larger than the median at the same time.
		\textbf{Best viewed in color with 300\% zoom}.
		}
	\label{fig_extreme}
\end{figure}
If we take derivative of $y^{nor}$ \wrt $\mu_y$,
we can easily find that $y^{nor}$ is decreasing on the interval $[0, 1]$.
Thus, $y^{nor}$ get the maximum $y^{nor}_{max}$
if and only if $y=1$ and values of all the other pixels are $0$.      
In the opposite case, $y^{nor}$ get the minimum $y^{nor}_{min}$. 
In practice, the $e_{max}$ in the Eq.~\eqref{equ_fnc} is set to
be the difference between the theoretical $y^{nor}_{max}$ and $y^{nor}_{min}$.
The distribution of the predicted probability $p$ is not so extreme like the
distribution of the ground truth, so the maximum of the absolute normalized probability
is usually smaller than the $\vert y^{nor}_{max} \vert$ or $\vert p^{nor}_{min} \vert$.
This is shown in Fig.~\ref{fig_extreme}.

As pointed by Wang \etal~\cite{2004_SSIM}, 
the statistical features of the image are usually highly spatially non-stationary.
Furthermore, the global mean and variance are rotation invariant.
To better capture the local details of the image,
we apply the SSL locally rather than globally.

Finally, we get the overall objective function:
\begin{align}
\mathcal{L}_{all}(y, p)= \lambda \mathcal{L}_{ce}(y, p) + (1 - \lambda)\mathcal{L}_{ssl}(y, p),
\end{align}
where the $\lambda \in [0, 1]$ is a weight factor and we simply set $\lambda=0.5$ in practice.
The role of the pixel-wise cross entropy loss $\mathcal{L}_{ce}$ is like the luminance
term (Eq.~\eqref{equ_002}) of the SSIM index.
They both measure the similarity of the pixel intensity between two images.
The role of the $\mathcal{L}_{ssl}$ is like the structure term (Eq.~\eqref{equ_004}).
They both measure the structural similarity between two images.

To be consistent with the SSL,
we use the sigmoid cross entropy loss rather than the multiclass softmax cross entropy loss.
It means we consider the semantic segmentation as
multiple one-vs.-rest binary classification problems
rather than a multi-class classification problem.
Thus we adopt the sigmoid cross entropy loss and train multiple binary classifiers jointly.
Liang \etal~\cite{2018_CVPR_DSSPN} argue
the class competition introduced by softmax cross entropy loss
hinders the model's capability of learning a unified model using diverse label annotations,
where only some parts of concepts belonging to one super-class are visible.
EncNet~\cite{2018_CVPR_EncNet} uses the sigmoid cross entropy loss to regularize the training.
In practice, we get a slight improvement
with the sigmoid cross entropy loss on the PASCAL VOC 2012 dataset~\cite{2010_PASCAL},
but the loss does not work on the Cityscapes dataset~\cite{2016_Cityscapes}.

\section{Experiments}

\subsection{Experimental setup} \label{sec_4_1}
\textbf{Models.}
We choose the DeepLabv3~\cite{chenDeepLabv3} and DeepLabv3+~\cite{chenDeepLabv3plus} as our base models.
The backbone networks of the DeepLabv3 and DeepLabv3+ models are ResNet-101~\cite{2016_ResNet}
and Xception-65~\cite{2017_Xception} respectively.
The backbone networks are pretrained on
ImageNet~\cite{ImageNet}\footnote{\url{https://github.com/tensorflow/models/blob/master/research/deeplab/g3doc/model_zoo.md}}.

\textbf{Datasets.} 
We evaluate our method on the PASCAL VOC 2012~\cite{2010_PASCAL} and Cityscapes~\cite{2016_Cityscapes} datasets.
The original PASCAL VOC 2012 dataset contains 1\,464~(\textit{train}), 1\,449~(\textit{val}) and 1\,456~(\textit{test}) images.
And the dataset contains 20 foreground object classes and one background class. 
Cityscapes dataset is a large-scale dataset
containing high quality pixel-level annotations
of 5\,000 images (2\,975, 500, and 1\,525 for the training, validation, test sets respectively).
Cityscapes dataset contains 19 object classes.

\begin{table*}[t]
	\centering
	\resizebox{0.96\textwidth}{12mm}{	
		\setlength{\tabcolsep}{0.6mm}{
			\begin{tabular}{l||ccccccccccccccccccccc||c}
				\toprule
				Method
				& backg.  		& aero. 		& bike 		& bird 
				& boat 			& bottle 		& bus 		& car 
				& cat 			& chair 		& cow 		& d.table 
				& dog 			& horse 		& mbike 	& person
				& p.plant 		& sheep 		& sofa 		& train 
				& tv 		& mIoU (\%) \\
				\midrule
				CE 			
				& 94.07			& 85.54				& 52.46 			& 81.18 
				& 67.84 		& 77.63 			& 91.42 			& 88.00
				& 91.27 		& 35.95 			& 83.44 			& 61.47 
				& 86.80 		& 84.48 			& 85.67 			& 82.28 
				& 64.19 		& 85.97 			& 53.63 			& 84.02
				& 69.97 		& 76.54		\\
				\midrule
				\rowcolor{mygray}
				SSL(3) 			
				& \textbf{94.43}& \textbf{88.58}	& \textbf{56.32} 	& \textbf{85.52} 
				& 67.22 		& 76.07 			& 90.53 			& 85.92
				& 90.68 		& 35.92 			& 81.80 			& \textbf{62.76} 
				& 86.80 		& \textbf{86.47} 	& \textbf{86.42} 	& \textbf{84.32}
				& \textbf{65.25}& \textbf{86.65} 	& \textbf{56.36} 	& 81.80
				& \textbf{72.55}& \textbf{77.26}		\\
				\midrule \midrule
				CE(+)
				& 94.54			& 91.96				& 58.00 			& 88.07 
				& 66.45 		& 78.17 			& 93.51 			& 88.36
				& 91.34			& 37.21 			& 84.90 			& 60.50 
				& 87.51 		& 82.06 			& 84.64				& 84.21 
				& 60.79 		& 84.76 			& 56.45 			& 87.59
				& 73.70 		& 77.84		\\
				\midrule
				\rowcolor{mygray}
				SSL(+3) 			
				& \textbf{95.25}& \textbf{93.38}	& \textbf{63.91} 	& 82.46
				& \textbf{70.83}& 76.47 			& \textbf{94.75} 	& \textbf{88.79}
				& \textbf{93.83}& 35.74 			& \textbf{87.04} 	& \textbf{62.35} 
				& \textbf{88.99}& \textbf{88.69} 	& \textbf{87.00} 	& \textbf{86.80} 
				& \textbf{64.39}& \textbf{88.47} 	& \textbf{60.25} 	& \textbf{90.24}
				& 70.98			& \textbf{79.55}	\\
				\bottomrule	
	\end{tabular}}}
	\caption{Per-class results on the PASCAL VOC 2012 \textit{test} set.
		The CE means the base model is DeepLabv3 and the loss is softmax cross entropy. 
		The SSL(+3) indicates that the base model is DeepLabv3+ and the loss is the SSL with region size 3.
	}
	\label{tab_003}
\end{table*}
\textbf{Learning rate and training steps.}
We use the poly learning rate policy where the initial learning rate is
multiplied by $(1 - \frac{\textit{iter}}{\textit{max\_iter}})^{\textit{power}}$ with $\textit{power}=0.9$.
For the PASCAL VOC 2012 dataset, the model is first trained on the
\textit{trainaug}~\cite{2011_PASCALAUG} set which contains 10\,582
images for about $30K$ iterations with the initial learning rate = 0.007,
then we freeze the batch normalization parameters and 
finetune the model on the \textit{train} set
for another $20K$ iterations with a smaller initial learning rate~=~0.0005.
For Cityscapes dataset, we train the model on the train set
for about 90K iterations with the initial learning rate = 0.007 and no finetuning procedure. 
We adopt the slow start strategy~\cite{2013_ICML_slow},
where a small learning rate is used for the beginning 100 steps,
0.001 for the first stage and 0.0001 for the finetune stage.

\textbf{Crop size and output stride.}
For the PASCAL VOC 2012 dataset,
when using DeepLabv3 as the base model,
the crop size is 513, and the batch size is 12.
With DeepLabv3+, we change the crop size to be 393. 
For the Cityscapes dataset,
when using DeepLabv3 as the base model,
the crop size is 669, and the batch size is 8.
We choose small crop sizes and batch sizes due to the limitations of our experiment platforms,
two GTX 1080 Ti~(11GB memory) GPUs.
As the DeepLabv3+ model adds a decoder module to the DeepLab3 model,
it needs more memory to hold the network.
A larger crop size or batch size can get better performance~\cite{chenDeepLabv3}.
The output stride is always 16 during training and inference.
It is also worth noting that
we upscale the logits back to the input image resolution
rather than downsampling the ground truth labels for training.

\textbf{Data augmentation.}
We apply data augmentation by randomly scaling the input 
images~(from 0.5 to 2.0 on PASCAL, 0.75 to 1.25 on Cityscapes) and randomly left-right flipping during training. 
After training, we do inference on the original images without any special inference strategy.
The evaluation metric is the mean intersection-over-union (mIoU) score.  
The rest settings are the same as~\cite{chenDeepLabv3}.

\subsection{Results on PASCAL VOC 2012 dataset}
\subsubsection{Effectiveness of the SSL}
\begin{table}[t]
	\centering
	\resizebox{0.43\textwidth}{!}{%
	\begin{tabular}{llll}  
		\toprule
		Model 		&Method  			 	&Size			&  mIoU~(\%)  		\\
		\midrule
		\multirow{12}{*}{DeepLabv3} 
		&CE  			&--	  		& 76.33 				\\
		~	&BCE			&--	    	& 76.80   			\\
		\cmidrule{2-4}
		~	&SSIM - Eq.~\eqref{equ_ssim_loss_lucs}	
							&3	    	& 77.16   			\\
		~	&SSIM - Eq.~\eqref{equ_ssim_loss_lucs}	
							&11	    	& 76.25    			\\
		~	&SSIM - Eq.~\eqref{equ_ssim_loss_cs}
							&3	    	& 77.02   			\\
		~	&SSIM - Eq.~\eqref{equ_ssim_loss_cs}
							&11	    	& 76.78   			\\
		\cmidrule{2-4}
		~	&SSL			&1			& 76.66 			\\
		~	&SSL			&11			& 77.70 			\\
		~	&SSL			&9			& 77.87 			\\
		~	&SSL			&7			& 78.34 			\\
		~	&SSL			&5			& 77.69 			\\
		~	&SSL			&3			& \textbf{78.46} 	\\
		\midrule[0.5mm]
		\multirow{6}{*}{DeepLabv3+}
		&CE 			&--	  			& 78.12  			\\
		~	&BCE			&--    		& 78.31   			\\
		\cmidrule{2-4}
		~	&SSL			&1			& 79.04 			\\
		~	&SSL			&11			& 79.22 			\\
		~	&SSL			&7			& 79.88 			\\		
		~	&SSL			&3			& \textbf{80.63} 	\\
		\bottomrule
	\end{tabular}}
	\caption{Evaluation of the SSL on the PASCAL VOC 2012 \textit{val} set.
		CE and BCE are the softmax and sigmoid cross entropy losses respectively.
		The size is the region size.
		When using the SSIM, we apply the sigmoid operation.}
	\label{tab_001}
\end{table}
We first evaluate the SSL on the PASCAL VOC 2012 \textit{val} set.
The results are shown in Tab.~\ref{tab_001}.
With DeepLabv3 and DeepLabv3+ as base models,
the SSL can improve the performance by 2.13\% and
2.51\% on the \textit{val} set respectively.
With similar settings, the DeepLabv3 and DeepLabv3+
can achieve 77.21\% and 78.85\% on the \textit{val} set respectively
in the original papers~\cite{chenDeepLabv3, chenDeepLabv3plus}.
They set the batch size to be 16, crop size to be 513,
and the models are only trained on the \textit{trainaug} set.
Our batch size is 12, and the crop size is 513 or 393.

As shown in Tab.~\ref{tab_001},
the performance of SSL with regions size 3 or 7 is obviously better than the 
performance of SSL with region size 1.
When the region size is 1, the SSL becomes a pixel-wise loss.
It demonstrates the SSL does encode the local structure information of the objects.
Compared to the SSIM (Eq.~\eqref{equ_ssim_loss_lucs} and Eq.~\eqref{equ_ssim_loss_cs}),
the SSL shows better performance.
This is consistent with our analysis in Sec.~\ref{sec_3_1},
the SSIM index is not an appropriate optimization criterion for semantic segmentation.
From Tab.~\ref{tab_001}, we can also find that the SSL with a smaller region size
is more likely to get better performance.
This may due to two larger image patches are more likely
to have similar statistics at the late stage of the training,
and this leads to small differences between the two normalized patches.
Thus the training becomes inefficient.

The per-class results on \textit{test} set are shown in Tab.~\ref{tab_003},
and they are given by the official evaluation server\footnote{\url{http://host.robots.ox.ac.uk:8080/anonymous/P25MWR.html}}.
With DeepLabv3 and DeepLabv3+ as base models,
the SSL can improve the performance by 0.72\% and 1.71\% on the \textit{test} set respectively.

\begin{table}[t]
	\centering
	\resizebox{0.44\textwidth}{!}{%
		\begin{tabular}{llcccl}  
			\toprule
			Size  	& $\sigma$			&$\beta$		& OHEM  		&  Reweight 	&  mIoU (\%)  		\\
			\midrule[0.5mm]
			3		& 1.5			& 0.10			&\checkmark		& 				&  	77.66		\\
			3		& 1.5			& 0.10			&				& \checkmark	&  	77.61		\\
			\midrule[0.5mm]
			3		& 1.0			& 0.10			&\checkmark		& \checkmark	&  	77.41		\\
			3 		& 2.0			& 0.10			&\checkmark		& \checkmark	&	77.92 		\\
			3 		& 2.5			& 0.10			&\checkmark		& \checkmark	&	77.99 		\\
			3 		& 3.0			& 0.10			&\checkmark		& \checkmark	&	78.36 		\\	
			3 		& 3.5			& 0.10			&\checkmark		& \checkmark	&	77.89 		\\
			\midrule[0.5mm]
			3 		& 1.5			& 0.06			&\checkmark		& \checkmark	&	77.39 		\\		
			3 		& 1.5			& 0.08			&\checkmark		& \checkmark	&	77.84 		\\
			3 		& 1.5			& 0.09			&\checkmark		& \checkmark	&	77.88 		\\
			3 		& 1.5			& 0.11			&\checkmark		& \checkmark	& 	78.02		\\
			3 		& 1.5			& 0.12			&\checkmark		& \checkmark	&	78.09 		\\
			3 		& 1.5			& 0.14			&\checkmark		& \checkmark	&	77.22 		\\
			\midrule[0.5mm]
			3		& 1.5			& 0.10			&\checkmark		& \checkmark	&  	\textbf{78.46}		\\
			\bottomrule					
	\end{tabular}}
	\caption{The influence of different components of the SSL.
		The base model is DeepLabv3.
		The $\sigma$ is the standard deviation of the gaussian kernel.
		The larger the sigma is, the closer the gaussian distribution to the uniform distribution.}
	\label{tab_ablation}
\end{table}

\subsubsection{Ablation study}
Furthermore, we study the influence of the OHEM, the reweighting strategy, the standard deviation of the
gaussian kernel and the factor $\beta$ in Eq.~\eqref{equ_fnc} used to choose the hard examples.
When studying the influence of the OHEM, we set all the weights to be $1$.
When studying the effect of the reweighting strategy, we do not abandon the easy examples
and only reweight the cross entropy loss.
The results on PASCAL VOC 2012 \textit{val} set are shown in Tab.~\ref{tab_ablation}.

\begin{figure}[t]
	\centering
	\includegraphics[width=0.48\textwidth]{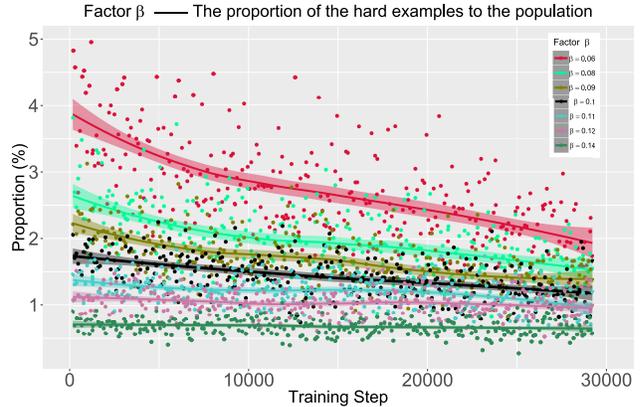}
	\caption{The proportion of the hard examples to the population
		with different $\beta$.
		The region size is 3, and the standard deviation of the gaussian kernel is 1.5.
		It is easy to see that the proportion is sensitive to the value of $\beta$.
		When $\beta=0.1$, the proportion is about $1.5\%$.
	\textbf{Best viewed in color with 300\% zoom}.
	}
	\label{fig_pro}
\end{figure}

From Tab.~\ref{tab_ablation}, we find both OHEM and reweighting strategies are essential.
When only applying one of the two strategies, the performance cannot obtain a high improvement.
And when the standard deviation of the gaussian kernel is 1.5, the SSL gets the best result.
We further record the proportion of the hard examples to the population with different $\beta$,
and this is shown in Fig.~\ref{fig_pro}.
It is clear that the number of hard examples is sensitive to the value of $\beta$,
so the choice of the $\beta$ is vital to the SSL.
The model gets the best performance when $\beta=0.01$.

\subsubsection{Comparison to prior work}

In this section, we compare the SSL with the CRF~\cite{2011_Efficient_CRF} and 
the affinity field loss~\cite{2018_ECCV_AAF} based on the DeepLabv3 model. 
We also show the results of GAN~\cite{2016_pauline}.

\textbf{CRF.}
Following the DeepLabv2~\cite{chenDeepLabv2}, we use the fully connected CRF
introduced in~\cite{2011_Efficient_CRF} as a post-processing step
and use the negative logarithm of the predicted probability as unary potential.
We use the default setting of the official public
code\footnote{\url{http://www.philkr.net/2011/12/01/nips/}}
and its python wrapper\footnote{\url{https://github.com/lucasb-eyer/pydensecrf}}.

As a post-processing step, CRF needs additional inference time.
When the inference steps of the CRF are 1, 2, and 5,
the additional time for each image is about 0.4s, 0.5s and 0.75s respectively.
This is intolerable in some real-time application scenarios.  

\textbf{Affinity field loss.}
The affinity field loss~\cite{2018_ECCV_AAF} exploits the relationships between pairs of pixels.
The loss imposes a grouping force on the neighbour pixels which belong to the same class
to make their predictions more consistent,
and a separating force on the neighbour pixels which belong to different classes
to make their predictions more inconsistent.

We reproduce the affinity field loss according to the official
implementation\footnote{\url{https://github.com/twke18/Adaptive\_Affinity\_Fields}}.
However, the loss adopts an 8-neighbour strategy and
we need to hold $8\times$ ground truth and predicted maps in memory when calculating the loss.
For PASCAL VOC dataset which contains 21 object classes,
if the value of a pixel is 32bits long,
the size of two tensors with shape $12 \times 513 \times 513 \times 21 \times 8$ is about 3.95GB.
Thus the loss is memory-consuming.
Ke \etal~\cite{2018_ECCV_AAF} downsample the label
map and use 4 GTX Titan X GPUs (12 GB memory) to hold 16 images in a batch with crop size to be 480. 
With only two GTX 1080Ti GPUs, we set crop size to be 321 and batch size to be 16 for all methods.
Other settings are same as described in Sec.~\ref{sec_4_1}.
\begin{table*}[t]
	\centering
	\resizebox{0.94\textwidth}{8.2mm}{	
		\setlength{\tabcolsep}{0.6mm}{
			\begin{tabular}{l||ccccccccccccccccccc||c}
				\toprule
				Method 
				& road  		& swalk 	& build. 		& wall 				
				& fence 		& pole 		& tlight 		& tsign 		
				& veg. 			& terrain 	& sky 			& person 			
				& rider 		& car	 	& truck 		& bus 				
				& train			& mbike 	& bike 			& mIoU (\%) \\
				\midrule
				CE 			
				& 97.97				& 82.97				& 91.16 			& 44.91	
				& 58.15				& 54.19				& 65.25 			& 74.23 	
				& 91.36 			& 59.81 			& 93.73 			& 79.36
				& 62.34 			& 94.40 			& 71.32 			& 81.87
				& 60.97 			& 62.30 			& 74.52 			& 73.73			\\
				\midrule
				\rowcolor{mygray}
				SSL(3)	
				& 97.95  			& \textbf{83.20} 	& \textbf{91.56} 	& \textbf{53.14}
				& 58.07 			& \textbf{54.86} 	& \textbf{66.85} 	& \textbf{75.88}	
				& \textbf{91.68} 	& \textbf{61.55} 	& \textbf{93.81} 	& \textbf{79.77} 	
				& 62.17 			& \textbf{94.75} 	& \textbf{79.12}	& \textbf{83.31}			
				& \textbf{63.68} 	& 60.71 			& \textbf{75.21}	& \textbf{75.12}  	\\
				\bottomrule	
	\end{tabular}}}
	\caption{Per-class results of the SSL with DeepLabv3 on the Cityscapes \textit{val} set.
		SSL(3) indicates the region size is 3.
	}
	\label{tab_005}
\end{table*}

\begin{table}[t]
	\centering
	\resizebox{0.44\textwidth}{!}{%
		\begin{tabular}{llll}  
			\toprule
			Model& Method  			 		& Size  	&  mIoU (\%)  			\\
			\midrule
			\multirow{2}{*}{Luc \etal~\cite{2016_pauline}} 		&
			Base 							& -- 		& 	71.79  				\\	
			~&	LargeFOV 					& --		&	72.04 				\\
			\midrule
			\multirow{9}{*}{DeepLabv3} 		&
			CE 								& -- 		& 	74.83  				\\
			~&	BCE							& --		&  	75.24				\\
			~&	CE + CRF-1					& --		&  	75.65				\\
			~&	CE + CRF-2					& --		&  	75.00				\\
			~&	CE + CRF-5					& --		&  	73.58				\\
			~&	Affinity					& 3			&  	75.29				\\
			\cmidrule{2-4}
			~&	SSL 						& 11		&	75.89 				\\
			~&	SSL 						& 7			&	76.08 				\\
			~&	SSL 						& 3			&	\textbf{76.76} 		\\
			\bottomrule					
	\end{tabular}}
	\caption{Comparison to prior work on PASCAL VOC 2012 \textit{val} set.
		Luc \etal~\cite{2016_pauline} use the base as their segmentation network,
		and the LargeFOV is an additional network which is used as the discriminator
		when applying the adversarial training strategy.
		When using the DeepLabv3 as the base model, the crop size is $321 \times 321$.
		CRF-X means that we do inference with X iteration steps when applying CRF.
		The size of the affinity field loss is the neighbour size used to choose the pixel pairs,
		and 3 is the default choice in ~\cite{2018_ECCV_AAF}.
		The SSL gets better performance than affinity field loss and CRF.
		Meanwhile, the SSL requires less additional memory and has no extra inference time.}
	\label{tab_006}
\end{table}
\textbf{GAN.}
Luc \etal~\cite{2016_pauline} get only about 0.25\%
improvement on the PASCAL VOC 2012 \textit{val} set,
and this is not effictive enough.
Additionally, the pix2pix~\cite{2017_Isola} and CycleGAN~\cite{2017_ICCV_CycleGAN}
can only achieve about 35.00\% and 16.00\% mIoU score on the Cityscapes \textit{val} set respectively.
Limited by the GPU memory, it is hard for us to add an
additional deep discriminator network to the DeepLabv3 model.
So we directly reference the experimental data from the paper~\cite{2016_pauline}.

\begin{table}[t]
	\centering
	\resizebox{0.40\textwidth}{!}{%
		\begin{tabular}{llll}  
			\toprule
			Model& Method  			 		& size  	&  mIoU~(\%) 		\\
			\midrule
			\multirow{7}{*}{DeepLabv3} 		&
			CE 								& -- 			& 	73.73  			\\
			~&	BCE 						& -- 			& 	73.58  			\\
			\cmidrule{2-4}
			~&	SSL							& 11			&  	74.52			\\
			~&	SSL							& 9				&  	74.29			\\
			~&	SSL							& 7				&  	74.26			\\
			~&	SSL							& 5				&  	74.27			\\
			~&	SSL							& 3				&  	\textbf{75.12}	\\	
			\bottomrule					
	\end{tabular}}
	\caption{Results of the SSL on Cityscapes \textit{val} set.
	}
	\label{tab_004}
\end{table}
All experimental results are shown in Tab.~\ref{tab_006}.
The CRF, affinity field loss and SSL can improve the performance of the model by 0.82\%, 0.46\%, and \textbf{1.93}\% respectively.
Compared to the CRF and affinity field loss, the SSL achieves better performance.
Meanwhile, the SSL is less time-consuming than CRF and less memory-consuming than the affinity field loss.
Compared to the GAN, the SSL requires no additional networks,
and it is evident that the SSL is more effective.
With these appealing properties,
any segmentation frameworks can incorporate the SSL effortlessly.

\subsection{Results on Cityscapes dataset}
Moreover, we evaluate the SSL
on the Cityscapes dataset with DeepLabv3 model.
The size of the image in Cityscapes dataset is $2048 \times 1024$.
As we can only apply the crop size to be about 489 
and batch size to be 8 with the DeepLabv3+ model on two GTX 1080 Ti GPUs,
the variation between the mini-batches is large, and this makes the training unstable. 
Thus we do not evaluate our methods with the DeepLabv3+ model on the Cityscapes.

The results are shown in Tab.~\ref{tab_004}.
With DeepLabv3 as the base model,
the SSL can improve the performance by 1.39\% on the Cityscapes \textit{val} set.
The per-class results are shown in Tab.~\ref{tab_005}.
With similar settings, the DeepLabv3 can achieve 77.23\% mIoU on the \textit{val} set
in the original paper~\cite{chenDeepLabv3}.
They set the batch size to be 16, and the crop size to be 769.
Our batch size is 8, and the crop size is 669.

\section{Conclusion}
In this work, we propose a correlation
maximized structural similarity loss for semantic segmentation.
We use the linear correlation between the ground truth map
and the predicted map to quantify their structural similarity.
And the goal of the SSL is to pay more attention to predictions which
lead to a low degree of linear correlation between the two maps.
Thus the model can achieve a strong structural similarity between the two maps by minimizing the SSL.
The SSL is simple yet effective and only requires a few additional computational resources during the training stage.
Moreover, we experimentally demonstrate that our method
can achieve substantial and consistent improvements in performance
on standard benchmark datasets.

%

\clearpage
\section*{Acknowledgments}

This work was supported in part by The National Key Research and Development Program of China (Grant Nos: 2018AAA0101400), in part by The National Nature Science Foundation of China (Grant Nos: 61936006), in part by the Alibaba-Zhejiang University Joint Institute of Frontier Technologies.

{
	
\bibliographystyle{ieee}
\bibliography{egbib}
}

%
%

\end{document}